\documentclass{article}

\PassOptionsToPackage{numbers, compress}{natbib}

\usepackage[preprint]{neurips_2025}
 \usepackage{amsmath}
\usepackage{fontawesome5}
\usepackage{enumitem}
\usepackage{wrapfig}
\usepackage[table,xcdraw]{xcolor}
\usepackage{multirow}
\usepackage[utf8]{inputenc} 
\usepackage[T1]{fontenc}    
\usepackage{hyperref}       
\usepackage{url}            
\usepackage{booktabs}       
\usepackage{amsfonts}       
\usepackage{nicefrac}       
\usepackage{microtype}      
\usepackage{xcolor}         
\usepackage[most]{tcolorbox}
\usepackage{fontawesome5}
\usepackage{hyperref}
\usepackage{array} 
\usepackage{listings}
\usepackage{graphicx}  
\usepackage{subcaption} 
\usepackage{float}      
\usepackage{fontawesome5}
\usepackage{booktabs}
\usepackage{graphicx}
\usepackage{colortbl}
\usepackage[table]{xcolor}
\usepackage{bbm}

\usepackage{caption}
\usepackage{capt-of}

\definecolor{avgblue}{RGB}{241,247,255}

\definecolor{lightblue}{RGB}{235,245,255} 
\definecolor{liautoblue}{RGB}{71,111,182} 
\definecolor{textred}{RGB}{128,0,0}
\usepackage{titlesec}
\titleformat{\section}
  {\large\bfseries\color{liautoblue}}{\thesection}{1em}{}
\titleformat{\subsection}
  {\bfseries\color{liautoblue}}{\thesubsection}{1em}{}
  \titleformat{\subsubsection}
  {\bfseries\color{liautoblue}}{\thesubsubsection}{1em}{}
\usepackage[most]{tcolorbox}
\newtcolorbox{liautoabstract}{
    colback=lightblue,
    colframe=white,
    boxrule=0pt,
    arc=2mm,
    left=4mm,
    right=4mm,
    top=5mm,
    bottom=5mm,
    enhanced, 
    before upper={\setlength{\parindent}{0pt}} 
}
\usepackage[most]{tcolorbox} 

\newtcolorbox{stepTitle}[1]{
    enhanced,
    colback=gray!5,    
    colframe=black!50,  
    boxrule=-1pt,
    arc=0mm,            
    left=2mm, right=2mm, top=1mm, bottom=1mm,
    fontupper=\small,  
    title=#1
}

\tcbset{
    stepbox/.style={
        enhanced,
        colback=gray!20,         
        colframe=gray!50,        
        boxrule=0.5pt,
        title={#1},              
        fonttitle=\bfseries,     
        left=2mm, right=2mm, top=1mm, bottom=1mm
    }
}

\newtcolorbox[auto counter]{case}[2][]{ 
    enhanced,
    colback=gray!5,
    colframe=black!70,
    coltitle=white,
    fonttitle=\bfseries\sffamily,
    fontupper=\small,
    arc=1.5mm,
    boxrule=0.5pt,
    title=Case study \thetcbcounter: #2, 
    left=1mm, right=1mm, top=2mm, bottom=2mm,
    label type=case, 
    #1               
}

\newtcolorbox{toolbox}[1]{
    enhanced,                 
    colback=gray!5,           
    colframe=black!70,        
    coltitle=white,           
    fonttitle=\bfseries\sffamily,
    fontupper=\small,
    arc=1.5mm,                
    boxrule=0.5pt,            
    title=#1,                 
    left=1mm, right=1mm, top=2mm, bottom=2mm
}

\usepackage{fancyhdr}
\usepackage{graphicx} 
\hypersetup{
    colorlinks=true,
    urlcolor=liautoblue,
}

\title{\Large LiAuto-GeoX: Efficient Grounded Driving Transformer}

\author{%
  Jiawei Lian$^{1,2}$\thanks{Equal Contribution} \And
  Haoyi Sun$^{2}$\footnotemark[1] \And
  Yang Wu$^{1}$\footnotemark[1] \And
  Lifu Mu$^{2}$\footnotemark[1] \And
  Siyuan Wang$^{1,2}$ \And
  Le Hui$^{3,4}$\thanks{Corresponding Author} \And
  Ning Mao$^{2}$\footnotemark[2]\, \thanks{Project Leader} \And
  Tao Wei$^{2}$ \And
  Pan Zhou$^{2}$ \And
  Kun Zhan$^{2}$ \And
  Jian Yang$^{1}$\footnotemark[2] \And
  \\[-0.5em]
  \centerline{\small $^{1}$Nanjing University of Science and Technology $^{2}$Li Auto Inc.
   $^{3}$Northwestern Polytechnical University}\\ \centerline{\small$^{4}$Department of Computing, The Hong Kong Polytechnic University}
}

\begin{document}

\maketitle

\begin{liautoabstract} 
Dense 3D reconstruction has demonstrated immense potential for spatial understanding, yet its viability as a real-time, onboard representation for autonomous driving remains an open challenge. Existing large-scale visual geometry models typically require substantial computational resources and lack the long-range geometric fidelity, surround-view consistency, and real-time efficiency demanded by dynamic driving environments.
To bridge this gap, we present \textbf{LiAuto-GeoX}, an efficient grounded driving transformer designed for deployable, ego-centric 3D scene understanding. Our approach begins by learning a high-capacity driving geometry model from large-scale surround-view data, utilizing sparse LiDAR priors to provide robust geometric grounding in distant, ambiguous, or structure-sparse regions. We then instantiate this capability into a highly compact 155M-parameter onboard model through a novel geometry-preserving distillation framework. This framework employs mask-guided depth-aware distillation to retain fine-grained metric structures by emphasizing geometrically informative regions, and relative-pose relational distillation to enforce cross-view spatial consistency through pose-induced geometric relations.
Extensive evaluations reveal that \textbf{LiAuto-GeoX} runs at 220 FPS on KITTI while maintaining high-fidelity dense reconstruction, enabling real-time deployment. The learned geometry transfers seamlessly to downstream autonomy tasks, achieving 90.6 PDMS in trajectory prediction, 24.63 mIoU in occupancy prediction, and 47.67 IoU in future-frame prediction. These all demonstrate that efficient dense 3D reconstruction can transcend its traditional role as a perception target to serve as a scalable, foundational geometric representation for next-generation autonomous driving.

\vspace{3mm}
    {\color{liautoblue!30}\rule{\linewidth}{0.5pt}} 
    \vspace{2mm}

    \small 
    \renewcommand{\arraystretch}{1.3} 
\begin{tabular}{@{} l l @{}}
        {\faCalendar*} & \textbf{Last Update Date:} June 1, 2026 \\
        {\color{liautoblue}\faGithub} & \textbf{Project:} {https://ljwwwiop.github.io/GeoX/} \\
    \end{tabular}\end{liautoabstract}

\section{Introduction}

%
While dense 3D reconstruction has long been a fundamental problem in computer vision, its role within autonomous driving is undergoing a paradigm shift.
For driving systems, 3D geometry transcends being merely a reconstruction target. 
Rather, it serves as a foundational representation for reasoning about free space, object layouts, dynamic agents, and future scene evolution from an ego-centric perspective~\cite{li2024bevformer,huang2023tri,wei2023surroundocc,hu2023planning, wu2026gem, zhang2024copilot4d}. 
Such a foundational role imposes strict practical constraints. An offline reconstruction model may yield highly accurate geometry, yet a practical onboard representation demands real-time efficiency, spatial consistency, and seamless integration with downstream autonomy tasks.
Consequently, the central question is no longer merely whether 2D images can be lifted to dense 3D, but whether this dense visual geometry can function as a deployable, foundational representation for real-world driving.

%
The rapid emergence of large-scale visual geometry models has made this inquiry increasingly pertinent.
From classical learning-based multi-view stereo to recent Transformer-based geometry architectures, visual reconstruction has evolved toward recovering metric 3D structure directly from images
~\cite{yao2018mvsnet,wang2024dust3r,murai2025mast3r,wang2025continuous,wang2025vggt,wang2025pi,zuo2025dvgt}. 
%
%
However, transferring these generic foundation models to autonomous driving poses non-trivial challenges.
%
%
On the spatial front, driving scenes rely on surround-view cameras with limited overlap, leading to sparse and uneven cross-view correlations. 
On the deployment front, autonomous vehicles impose strict deployment constraints, requiring perception modules to operate with low-latency inference under restricted onboard computation. 
Consequently, despite their impressive reconstruction capabilities, existing large-scale geometry models have yet to be established as a viable, deployable substrate for driving intelligence.

\begin{figure*}[t] 
  \centering 
  \includegraphics[width=0.98 \textwidth]{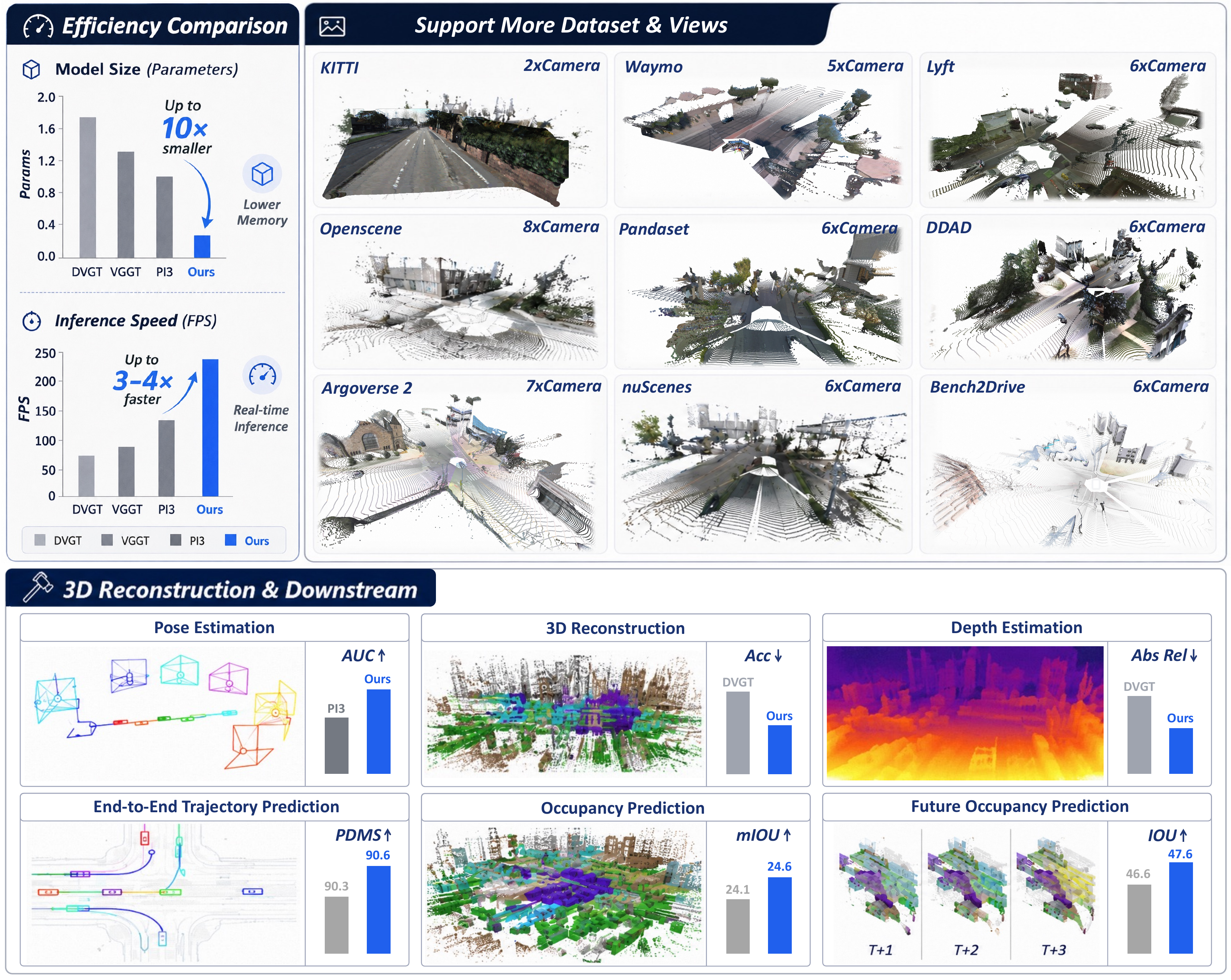} 
    \caption{\textbf{Overview of LiAuto-GeoX.}
    LiAuto-GeoX provides an efficient driving geometry model for surround-view 3D reconstruction and downstream autonomy tasks. 
    It strikes a compact yet effective accuracy–efficiency balance, accommodates diverse camera configurations, and transfers the learned dense geometry to tasks such as pose estimation, depth estimation, 3D reconstruction, trajectory prediction, occupancy prediction, and future occupancy forecasting, achieving excellent performance across all of them.
    }
  \label{fig:teaserfig1} 
  \vspace{-0.5cm}
\end{figure*}

A natural direction is to transfer the capability of a high-capacity geometry model into a compact onboard model. Yet this transfer cannot be treated as conventional model compression. 
Dense driving geometry is highly sensitive to both local and global errors. 
Locally, inaccurate depth around object boundaries, thin structures, or distant regions can lead to distorted geometry, especially under complex weather and lighting conditions~\cite{wu2025weathergen, wang2021regularizing}. 
Globally, inconsistent spatial relations across surround-view cameras can corrupt the ego-centric scene layout, which is particularly evident when the camera configurations or quantity varies~\cite{wang2025vggt,wang2025pi,zuo2025dvgt}.
%
%
Standard distillation objectives, such as generic feature matching or logits alignment~\cite{hinton2015distilling,yang2022masked,zhao2022decoupled}, fall short in two critical aspects: they treat all spatial regions uniformly and model multiple views independently. Under large teacher-student capacity gaps, a compact student tends to fit dominant, easy regions while losing fine-grained depth at boundaries, and fails to maintain globally consistent spatial relations across cameras with limited overlap. 
This critical gap motivates our geometry-preserving transfer framework that explicitly maintains fine-grained depth fidelity and cross-view spatial consistency.

In this work, we present \textbf{LiAuto-GeoX}, an efficient grounded driving transformer for deployable 3D driving scene understanding.
We first learn a high-capacity driving geometry teacher from large-scale surround-view data.
Crucially, instead of relying on visual appearance alone, we condition the teacher on the vehicle's calibrated multi-camera poses as explicit geometric inputs.
This pose-conditioned formulation anchors visual observations in a drive-centric coordinate system and enables the teacher to predict dense driving geometry with stronger metric consistency, especially in distant, ambiguous, or weakly textured regions.
We then instantiate this capability in a compact 155M-parameter onboard model through geometry-preserving distillation.
Specifically, Mask-Guided Depth-Aware Distillation emphasizes reliable and geometrically informative regions to preserve fine-grained metric structure, while Relative-Pose Relational Distillation transfers pose-induced cross-view relations to maintain globally consistent driving geometry.
To provide stronger student supervision, we distill dense predictions from the high-capacity teacher rather than sparse LiDAR measurements, yielding more continuous geometry signals for distant, ambiguous, and structure-sparse regions.

The resulting model achieves a strong balance between reconstruction fidelity and deployment efficiency. LiAuto-GeoX runs at 220 FPS on KITTI while retaining high-quality dense 3D reconstruction. More importantly, the learned geometry is not limited to reconstruction itself. It transfers effectively to downstream autonomous driving tasks, achieving 90.6 PDMS for trajectory prediction, 24.63 Occ mIoU for occupancy prediction, and 47.67 IoU for future-frame prediction. These results show that efficient dense 3D reconstruction can serve not only as a perception objective, but also as a scalable geometric representation for next-generation autonomous driving systems.

Our contributions are summarized as follows:
\begin{itemize}
    \item \textbf{Deployable driving geometry.} We introduce LiAuto-GeoX, an Efficient Grounded Driving Transformer that explores dense 3D reconstruction as a practical onboard representation for autonomous driving.

    \item \textbf{Geometry-preserving transfer.} We propose a geometry-preserving transfer framework that adapts large-scale driving geometry models into compact deployable models while retaining local depth awareness and cross-view consistency.

    \item \textbf{Real-time performance and task transfer.} LiAuto-GeoX runs at 220 FPS on KITTI video depth and shows effective transfer to trajectory prediction, occupancy prediction, and future-frame prediction.
\end{itemize}

\section{Related Work}

\textbf{3D Representations for Autonomous Driving.}
A central problem in autonomous driving is to construct a spatial representation that supports perception, prediction, and planning. Early vision-centric methods commonly lift image features into a bird's-eye-view (BEV) space, providing a unified coordinate system for multi-camera reasoning. Representative works such as Lift-Splat-Shoot~\cite{philion2020lift}, BEVDet~\cite{huang2021bevdet}, BEVDet4D~\cite{huang2022bevdet4d}, BEVFormer~\cite{li2024bevformer}, BEVDepth~\cite{li2023bevdepth}, and BEVFusion~\cite{liang2022bevfusion} have shown that BEV representations are effective for camera-based 3D detection, map segmentation, temporal fusion, and multi-sensor fusion. These methods significantly advance vision-centric autonomous driving by turning perspective-view features into task-friendly driving representations.


More recently, 3D occupancy prediction has emerged as a denser representation for scene understanding. TPVFormer~\cite{huang2023tri} extends BEV to tri-perspective views to better encode vertical structure, while VoxFormer~\cite{li2023voxformer} and SurroundOcc~\cite{wei2023surroundocc} predict volumetric occupancy from camera inputs. Benchmarks and datasets such as OpenOccupancy~\cite{wang2023openoccupancy} and Occ3D~\cite{tian2023occ3d} further promote occupancy prediction as a general 3D perception task. To reduce the cost of dense voxel supervision or inference, RenderOcc~\cite{pan2024renderocc} explores 2D rendering supervision, and FlashOcc~\cite{yu2023flashocc} improves efficiency through a channel-to-height design.
Beyond static spatial perception, recent frameworks have expanded driving representations to integrate downstream reasoning and temporal forecasting. UniAD~\cite{hu2023planning} integrates perception, prediction, and planning in a unified framework, while DriveLM~\cite{sima2024drivelm} and DriveVLM~\cite{tian2024drivevlm} introduce vision-language reasoning into autonomous driving. Concurrently, an emerging line of research focuses on driving world models to forecast the future evolution of driving scenes. For instance, occupancy and LiDAR world models~\cite{gu2024dome,du2025sparseworld,shi2026come,wu2026gem,zhang2024copilot4d,lian2026} leverage volumetric occupancy or point cloud sequences to simulate dynamic future states, providing temporally coherent spatial priors that explicitly benefit planning and decision-making.

Despite these advances, most existing driving representations are optimized for task-specific objectives such as detection, occupancy, segmentation, or planning. BEV representations are efficient but compress 3D structure into a top-down plane, while voxel occupancy methods provide denser spatial descriptions but often suffer from high memory and computation costs. In contrast, our work studies dense 3D reconstruction itself as a deployable geometric representation for autonomous driving. LiAuto-GeoX directly recovers dense 3D structure from surround-view images and further validates the learned geometry on downstream autonomy tasks.

\textbf{Dense Visual Geometry and Driving Reconstruction.}
Dense 3D reconstruction has long been studied through neural scene representations such as NeRF~\cite{mildenhall2021nerf} and 3D Gaussian Splatting~\cite{kerbl20233d}, which achieve high-fidelity reconstruction and novel-view synthesis. However, these methods typically rely on per-scene optimization, dense view coverage, and iterative rendering, making them difficult to directly deploy in real-time autonomous driving systems.

Recent visual geometry models have shifted toward direct reconstruction from images with stronger generalization. Large Reconstruction Model~\cite{hong2024lrm} and Depth Anything~\cite{yang2024depthv1,yang2024depth} demonstrate the effectiveness of large-scale training for 3D-related prediction. DUSt3R~\cite{wang2024dust3r} formulates dense unconstrained stereo reconstruction as pointmap regression, and MASt3R~\cite{leroy2024grounding} further grounds image matching in 3D geometry. CUT3R~\cite{wang2025continuous} introduces a continuous 3D perception model with persistent state, while VGGT~\cite{wang2025vggt} directly predicts camera parameters, depth maps, point maps, and point tracks from multiple images. Pi3~\cite{wang2025pi} explores permutation-equivariant visual geometry learning to reduce dependence on fixed reference views. These methods show that dense visual geometry can be learned at scale and generalized across diverse scenes.

Autonomous driving imposes different constraints. Driving inputs come from surround-view cameras with limited overlap and sparse cross-view correlations, while onboard systems require low latency under limited computation. Recent driving-oriented models such as DVGT~\cite{zuo2025dvgt} adapt visual geometry reconstruction to driving scenes, while FastVGGT~\cite{shen2025fastvggt} and LiteVGGT~\cite{shu2025litevggt} improve the inference efficiency of large visual geometry transformers. Nevertheless, these models remain costly for real-time onboard deployment. Our work addresses this gap by learning robust driving geometry from large-scale surround-view data and transferring it to a compact 155M-parameter model for real-time dense reconstruction.

\textbf{Knowledge Distillation for 3D Perception.}
Knowledge distillation transfers knowledge from large teachers to smaller students through soft predictions, feature alignment, attention transfer, or masked objectives~\cite{hinton2015distilling,srivastava2015training,zagoruyko2016paying,zhao2022decoupled,yang2022masked}. In autonomous driving, distillation has been widely used for efficient 3D perception and deployment, including cross-modal knowledge distillation~\cite{chen2022bevdistill,wang2023distillbev,zhou2023unidistill,wu2026distill}, camera-based 3D detection~\cite{klingner2023x3kd}, target geometry transfer~\cite{huang2022tig}, and point cloud detection~\cite{zhang2023pointdistiller}. These methods mainly transfer semantic or task-specific representations, such as object-level features, BEV features, or LiDAR-to-camera cues.

In contrast, dense 3D reconstruction requires transferring continuous metric geometry, including depth fidelity, point-level structure, and cross-view consistency. Although recent feed-forward reconstruction models have shown strong geometric capability, how to distill such models into deployable students remains largely unexplored in driving scenarios. Existing 3D distillation methods are not designed for this setting, where long-range depth accuracy and multi-view spatial relations must be preserved under large model compression. LiAuto-GeoX addresses this gap with a geometry-preserving transfer framework: Mask-Guided Depth-Aware Distillation strengthens supervision for distant and geometrically reliable regions, while Relative-Pose Relational Distillation maintains cross-view geometric consistency.

\begin{figure*}[t] 
  \centering 
  \includegraphics[width=1.0 \textwidth]{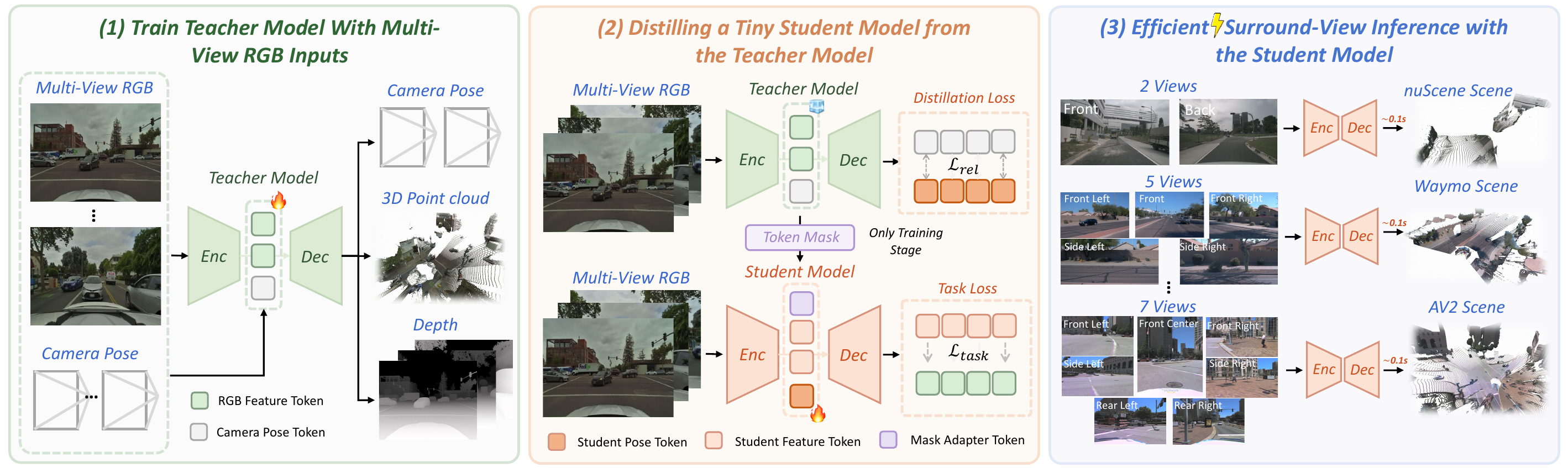} 
    \caption{\textbf{Overall pipeline of LiAuto-GeoX.}
    LiAuto-GeoX first trains a high-capacity teacher model to learn dense driving geometry from calibrated multi-view RGB inputs, then distills its geometric capability into a compact student model with task supervision, token-mask condition, and relational constraints. During inference, only the student is deployed to produce dense 3D reconstructions under flexible surround-view camera configurations.}
  \label{fig:framework} 
\end{figure*}

\section{Methodology}

\subsection{Overview}

We formulate \textbf{LiAuto-GeoX} as a teacher-student framework for real-time surround-view 3D reconstruction. Given calibrated surround-view images $\mathcal{I}=\{I_i\}_{i=1}^{N}$, where $I_i\in\mathbb{R}^{3\times H\times W}$, and camera parameters $\mathcal{C}=\{(K_i,T_i)\}_{i=1}^{N}$, the model predicts dense 3D driving geometry, including per-view depth maps $\mathcal{D}=\{D_i\}_{i=1}^{N}$ and a reconstructed point cloud $\mathcal{P}\in\mathbb{R}^{M\times 3}$. The overall pipeline, shown in Figure~\ref{fig:framework}, consists of three stages: teacher training, geometry-preserving distillation, and efficient surround-view inference. We first train a high-capacity driving geometry teacher $F_{\theta}^{T}$ from large-scale surround-view data, where sparse LiDAR priors provide geometric grounding in challenging regions. The teacher produces reliable reconstruction targets and intermediate geometry tokens, denoted as $(\mathcal{D}^{T},\mathcal{P}^{T},\mathcal{Z}^{T})$.

We then distill the teacher into a compact 155M-parameter student $F_{\theta}^{S}$ for real-time deployment. Instead of relying on generic output or feature alignment, LiAuto-GeoX transfers geometry through two complementary objectives: Mask-Guided Depth-Aware Distillation for reliable depth structure and Relative-Pose Relational Distillation for cross-view spatial consistency. After training, only the student is deployed, taking calibrated surround-view RGB images as input and producing dense 3D reconstructions in a single forward pass.

\subsection{Geometry-Aware Distillation}

\begin{figure*}[!t]
  \centering 
  \includegraphics[width=0.92 \textwidth]{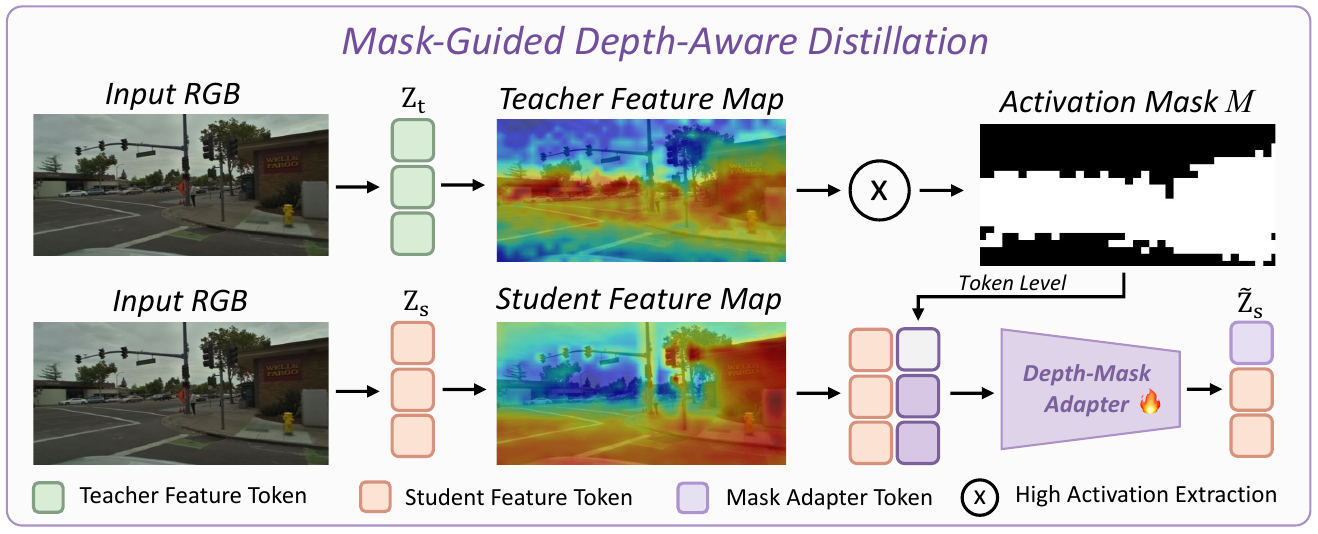} 
    \caption{\textbf{Mask-Guided Depth-Aware Distillation.}
    The frozen teacher model extracts high-activation regions to form a token-level mask, which conditions the student model through a lightweight depth-mask adapter. This training-time guidance focuses the compact student on geometrically informative regions without adding inference cost.}
  \label{fig:method_mask} 
\end{figure*}

To bridge the representational gap between the heavy teacher and the lightweight student, we propose a Geometry-Aware Distillation module. As illustrated in Figure~\ref{fig:method_mask}, this module consists of two core components designed to preserve critical 3D structures during cross-scale compression: Mask-Guided Depth-Aware Distillation and Relative-Pose Relational Distillation. Unlike direct feature alignment that treats all spatial regions and views independently, our approach explicitly enforces local geometric constraints and global multi-view structural consistency.

\subsubsection{Mask-Guided Depth-Aware Distillation}

A compact student often tends to fit dominant and easy regions, such as road surfaces or large static structures, while losing depth fidelity on distant objects, boundaries, and geometry-sensitive regions. This issue is particularly critical in driving scenes, where small depth errors in the far range can lead to large 3D localization deviations. Instead of introducing an additional masked distillation loss, we use the teacher's geometric response as a conditional signal to guide the student's feature formation. The key idea is to let the teacher identify informative depth regions and use the resulting mask to modulate the student tokens during training.

Let $\mathbf{Z}^{T}\in\mathbb{R}^{L\times C}$ and $\mathbf{Z}^{S}\in\mathbb{R}^{L\times C}$ denote the teacher and student tokens at a selected decoder layer, where $L$ is the number of spatial tokens and $C$ is the channel dimension. We first compute a token-level activation score from the frozen teacher:
\begin{equation}
    a_i = \left\|\mathrm{LN}(\mathbf{z}^{T}_{i})\right\|_2,\quad i=1,\dots,L,
\end{equation}
where $\mathbf{z}^{T}_{i}$ is the $i$-th teacher token and $\mathrm{LN}(\cdot)$ denotes layer normalization. A binary activation mask $\mathbf{M}\in\{0,1\}^{L}$ is then generated by selecting high-response teacher tokens:
\begin{equation}
    M_i = \mathbbm{1}(a_i \geq \tau),
\end{equation}
where $\tau$ is determined by the mean activation intensity. The mask is detached from the teacher branch and treated as a training-time condition, highlighting regions that carry stronger geometric responses.

To make this condition depth-aware, we downsample the teacher-predicted depth $\mathbf{D}^{T}$ to the token grid and normalize it as $\bar{\mathbf{d}}^{T}\in\mathbb{R}^{L}$. The activation mask and the depth signal are then encoded by a lightweight depth-mask adapter $\phi_{\psi}$:
\begin{equation}
    \mathbf{B}
    =
    \phi_{\psi}\left([\mathbf{M}, \bar{\mathbf{d}}^{T}]\right),
    \qquad
    \mathbf{B}\in\mathbb{R}^{L\times C},
\end{equation}
where $[\cdot,\cdot]$ denotes channel-wise concatenation. The adapter produces conditional bias tokens that are injected into the student representation:
\begin{equation}
    \widetilde{\mathbf{Z}}^{S}
    =
    \mathbf{Z}^{S}
    +
    \mathbf{M}\odot \mathbf{B},
\end{equation}
where $\odot$ denotes token-wise gating with broadcasting along the channel dimension. In this way, the teacher mask does not serve as a separate supervision target; instead, it conditions the student to allocate more representational capacity to reliable and depth-informative regions.

The conditioned tokens $\widetilde{\mathbf{Z}}^{S}$ are used by the following decoder layers to predict depth and 3D geometry, and are optimized through the overall training objectives described in Sec.~\ref{sec:total_loss}. Since the teacher branch is frozen and the mask is only used during training, this design introduces no additional inference cost. Compared with uniform feature alignment, mask-guided depth conditioning provides a more targeted transfer signal, improving the student's depth fidelity for distant objects, boundaries, and structure-sensitive regions while keeping the deployed model compact.

\subsubsection{Relative-Pose Relational Distillation}

\begin{figure*}[t] 
  \centering 
  \includegraphics[width=0.98 \textwidth]{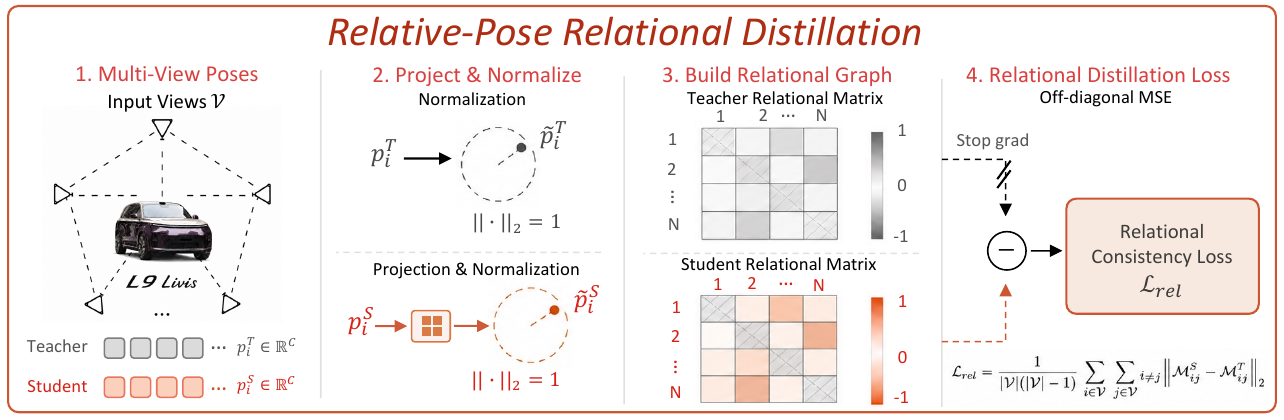} 
  \caption{\textbf{Relative-Pose Relational Distillation.}
    We normalize the teacher and student camera-conditioned tokens. We then construct pairwise relational matrices across input views to distill cross-view geometric relations. The student is supervised with an off-diagonal relational consistency loss, which transfers the teacher's view-to-view structure while avoiding direct absolute pose regression.}
  \label{fig:method_pose} 
\end{figure*}

Surround-view reconstruction requires the student to capture the geometric relations among cameras, especially when their image overlap is limited. As illustrated in Figure~\ref{fig:method_pose}, a straightforward strategy is to align each camera-conditioned token from the student to its teacher counterpart independently. However, such point-wise alignment ignores the relational structure among views, which is crucial for constructing a coherent 3D driving scene. We therefore introduce Relative-Pose Relational Distillation (RPR), which transfers the pairwise cross-view geometry encoded by the teacher to the student. Here, ``relative-pose'' refers to the relational structure among camera-conditioned tokens, rather than explicit ego-pose prediction.

Given a set of input views $\mathcal{V}=\{1,\dots,N\}$, we denote the teacher and student camera-conditioned tokens for the $i$-th view as $\mathbf{p}^{T}_{i}, \mathbf{p}^{S}_{i}\in\mathbb{R}^{C}$, respectively. These tokens encode the view-specific geometric context after interacting with image features and camera information. We first normalize each token onto a unit hypersphere:
\begin{equation}
    \widetilde{\mathbf{p}}^{T}_{i}
    =
    \frac{\mathbf{p}^{T}_{i}}{\|\mathbf{p}^{T}_{i}\|_2},
    \qquad
    \widetilde{\mathbf{p}}^{S}_{i}
    =
    \frac{\mathbf{p}^{S}_{i}}{\|\mathbf{p}^{S}_{i}\|_2}.
\end{equation}

We then construct teacher and student relational matrices by computing pairwise cosine similarities across all camera views:
\begin{equation}
    \mathbf{R}^{T}_{ij}
    =
    \left\langle
    \widetilde{\mathbf{p}}^{T}_{i},
    \widetilde{\mathbf{p}}^{T}_{j}
    \right\rangle,
    \qquad
    \mathbf{R}^{S}_{ij}
    =
    \left\langle
    \widetilde{\mathbf{p}}^{S}_{i},
    \widetilde{\mathbf{p}}^{S}_{j}
    \right\rangle .
\end{equation}
The relational matrix captures how different views are arranged in the learned geometric space. Compared with independent token regression, this formulation distills the structural relationship among cameras and is therefore better aligned with surround-view 3D reconstruction.

Since the diagonal terms correspond to self-similarity, we only optimize the off-diagonal entries:
\begin{equation}
    \mathcal{L}_{\mathrm{RPR}}
    =
    \frac{1}{N(N-1)}
    \sum_{i=1}^{N}
    \sum_{\substack{j=1 \\ j\neq i}}^{N}
    \left\|
    \mathbf{R}^{S}_{ij}
    -
    \mathrm{sg}(\mathbf{R}^{T}_{ij})
    \right\|_2^2 ,
\end{equation}
where $\mathrm{sg}(\cdot)$ denotes stop-gradient operation on the teacher branch.

This relational objective provides a global geometric constraint for the student. By matching the teacher's cross-view correlation structure, the student learns to preserve camera-to-camera spatial consistency without directly regressing absolute pose values. This is particularly important for driving scenes, where surround-view cameras often have limited overlap and sparse cross-view correspondences.

\subsection{Training Procedure and Loss Formulation}
\label{sec:total_loss}

LiAuto-GeoX is trained in two stages. We first optimize a high-capacity teacher to obtain reliable dense geometry supervision, and then freeze the teacher to train a compact student with geometry-preserving transfer. This design separates two objectives: the teacher focuses on constructing accurate driving geometry from large-scale surround-view data, while the student focuses on inheriting this capability under strict deployment constraints.

\paragraph{High-Capacity Geometry Learning.}
Autonomous driving platforms are typically equipped with rigidly mounted surround-view cameras, whose intrinsic and extrinsic parameters are calibrated and remain fixed with respect to the vehicle body. This property provides a reliable geometric prior that is often unavailable in generic image reconstruction. We therefore condition the high-capacity teacher on camera parameters during training, allowing it to learn driving-specific geometry under the calibrated multi-camera setting.

Given surround-view images $\mathcal{I}$ and camera parameters $\mathcal{C}$, the teacher $F_{\theta}^{T}$ predicts per-view depths, a dense 3D point cloud, and intermediate geometry tokens:
\begin{equation}
    (\mathcal{D}^{T}, \mathcal{P}^{T}, \mathcal{Z}^{T})
    =
    F_{\theta}^{T}(\mathcal{I}, \mathcal{C}),
\end{equation}
where $\mathcal{D}^{T}=\{D_i^{T}\}_{i=1}^{N}$ denotes the predicted depth maps, $\mathcal{P}^{T}\in\mathbb{R}^{M\times3}$ denotes the reconstructed 3D point cloud, and $\mathcal{Z}^{T}$ denotes the teacher geometry tokens. Let $\mathcal{D}^{*}$ and $\mathcal{P}^{*}$ be the available depth and point-cloud targets, with valid regions $\Omega_i^{D}$ and $\Omega^{P}$. The teacher reconstruction objective is:
\begin{equation}
\begin{aligned}
    \mathcal{L}_{\mathrm{rec}}^{T}
    =
    &\lambda_{D}
    \sum_{i=1}^{N}
    \frac{1}{|\Omega_i^{D}|}
    \sum_{\mathbf{u}\in\Omega_i^{D}}
    \rho\left(D_i^{T}(\mathbf{u})-D_i^{*}(\mathbf{u})\right) \\
    &+
    \lambda_{P}
    \frac{1}{|\Omega^{P}|}
    \sum_{\mathbf{x}\in\Omega^{P}}
    \rho\left(\mathcal{P}^{T}(\mathbf{x})-\mathcal{P}^{*}(\mathbf{x})\right),
\end{aligned}
\end{equation}
where $\rho(\cdot)$ is a robust regression loss.

To further improve the teacher in distant, ambiguous, and structure-sparse regions, we use sparse LiDAR only as a training-time geometric grounding signal. Specifically, LiDAR points are projected into each camera view to obtain sparse depth targets $D_i^{L}$ on valid pixels $\Omega_i^{L}$:
\begin{equation}
    \mathcal{L}_{\mathrm{lidar}}
    =
    \sum_{i=1}^{N}
    \frac{1}{|\Omega_i^{L}|}
    \sum_{\mathbf{u}\in\Omega_i^{L}}
    \rho\left(D_i^{T}(\mathbf{u})-D_i^{L}(\mathbf{u})\right).
\end{equation}
The teacher is optimized by:
\begin{equation}
    \mathcal{L}_{T}
    =
    \mathcal{L}_{\mathrm{rec}}^{T}
    +
    \lambda_{L}\mathcal{L}_{\mathrm{lidar}}.
\end{equation}
After convergence, the teacher is frozen and serves as a stable source of dense geometry targets, geometry tokens, and depth-aware mask conditions for student training.

\paragraph{Efficient Student Distillation.}
The student $F_{\theta}^{S}$ has 155M parameters and is trained to reproduce the teacher's dense geometry with substantially lower computation. During training, the teacher generates pseudo targets $(\mathcal{D}^{T},\mathcal{P}^{T})$ and depth-aware mask conditions. The mask-guided depth-aware module described above is used as a conditional adapter for student token formation; it does not introduce an additional loss term. The student prediction is written as:
\begin{equation}
    (\mathcal{D}^{S}, \mathcal{P}^{S}, \mathcal{Z}^{S})
    =
    F_{\theta}^{S}(\mathcal{I}, \mathcal{C}; \mathrm{sg}(\mathcal{M}^{T})),
\end{equation}
where $\mathcal{M}^{T}$ denotes the teacher-generated depth-aware activation masks and $\mathrm{sg}(\cdot)$ stops gradients through the teacher branch.

The student reconstruction objective follows the same form as the teacher objective, but uses the frozen teacher predictions as supervision:
\begin{equation}
\begin{aligned}
    \mathcal{L}_{\mathrm{rec}}^{S}
    =&
    \lambda_{D}
    \sum_{i=1}^{N}
    \frac{1}{|\Omega_i^{D}|}
    \sum_{\mathbf{u}\in\Omega_i^{D}}
    \rho\left(D_i^{S}(\mathbf{u})-\mathrm{sg}(D_i^{T}(\mathbf{u}))\right) \\
    &+
    \lambda_{P}
    \sum_{i=1}^{N}
    \frac{1}{|\Omega_i^{P}|}
    \sum_{\mathbf{u}\in\Omega_i^{P}}
    \rho\left(P_i^{S}(\mathbf{u})-\mathrm{sg}(P_i^{T}(\mathbf{u}))\right).
\end{aligned}
\end{equation}
To preserve global surround-view consistency, we add the relative-pose relational loss $\mathcal{L}_{\mathrm{RPR}}$ defined in the previous section. The final student objective is:
\begin{equation}
    \mathcal{L}_{S}
    =
    \mathcal{L}_{\mathrm{rec}}^{S}
    +
    \lambda_{\mathrm{rpr}}\mathcal{L}_{\mathrm{RPR}}.
\end{equation}
Thus, the student receives local depth-aware guidance through teacher-conditioned token adaptation, and global camera-relation supervision through $\mathcal{L}_{\mathrm{RPR}}$. During inference, the teacher, LiDAR inputs, and mask adapter are removed; only the compact student is deployed to produce dense 3D reconstruction in a single forward pass.

\begin{table*}[t]
\centering
\caption{\textbf{Comparison of the aggregated autonomous driving datasets.} We summarize the key sensor configurations and data modalities across the seven datasets used for our large-scale unified training. H-FOV denotes the Horizontal Field of View configuration for the camera setup.}
\label{tab:dataset_comparison}
\resizebox{\textwidth}{!}{
\begin{tabular}{l | c c c c c c c}
\toprule
\textbf{Attribute} & \textbf{Waymo} & \textbf{nuScenes} & \textbf{Pandaset} & \textbf{Lyft} & \textbf{DDAD} & \textbf{VKITTI} & \textbf{OpenScene} \\
\midrule
\textbf{Cameras} & 5 & 6 & 6 & 7 & 6 & 2 (Stereo) & 8 \\
\textbf{Frequency (Hz)} & 10 & 12 & 10 & 10 & 10 & 10 & 10 \\
\textbf{Resolution} & $1920 \times 1280$ & $1600 \times 900$ & $1920 \times 1080$ & $1224 \times 1024$ & $1936 \times 1216$ & $1242 \times 375$ & High-Res \\
\textbf{H-FOV Setup} & $50^\circ \times 5$ & $70^\circ \times 1 + 55^\circ \times 5$ & $360^\circ$ Coverage & $360^\circ$ Coverage & $120^\circ \times 6$ & $90^\circ \times 2$ & $360^\circ$ (8-Cam) \\
\textbf{Data Type} & Real & Real & Real & Real & Real & Synthetic & Real \\
\bottomrule
\end{tabular}
}
\end{table*}

\section{Experiments}

\textbf{Datasets.} To train our LiAuto-GeoX, we aggregate seven open-source autonomous driving benchmarks (Waymo~\cite{sun2020scalability}, nuScenes~\cite{caesar2020nuscenes}, Pandaset~\cite{xiao2021pandaset}, Lyft~\cite{houston2021one}, DDAD~\cite{guizilini20203d}, KITTI~\cite{geiger2012we}, and OpenScene~\cite{contributors2023openscene}) to construct a large-scale unified dataset, facilitating the generation of high-fidelity, dense 3D point clouds. 
Due to the sheer scale of OpenScene, we sample only 5K surround-view frames (8-camera setup) for joint training. 
As summarized in Table~\ref{tab:dataset_comparison}, this aggregated dataset exhibits significant diversity in sensor configurations, spatial distributions, and data modalities. 
Beyond demonstrating our model's efficacy in complex real-world scenarios on this dataset, we further perform direct inference on unseen datasets. These evaluations validate the exceptional transferability and robustness of our model in entirely novel environments.

\textbf{Implementation Details.}
For the teacher network, we adopt a large-scale visual geometry architecture following VGGT~\cite{wang2025vggt} and OmniVGGT~\cite{peng2025omnivggt}. The teacher consists of a 24-layer encoder and a 24-layer decoder with a feature dimension of 1024, providing high-capacity geometric supervision. 
For the deployable student, we use DINOv2-Small~\cite{oquab2023dinov2} as the encoder, which contains 12 Transformer layers, and pair it with a 12-layer decoder. The student feature dimension is reduced to 384, resulting in a compact model with 155M parameters. 
The whole framework is trained end-to-end on 32 NVIDIA A100 GPUs for approximately 10 days. We further adopt gradient checkpointing to reduce GPU memory consumption during training.

\textbf{Evaluation Metrics.} To comprehensively validate our model, we adopt standard evaluation protocols across multiple tasks. 
Notably, all validation and testing sets are constructed by randomly sampling 10\%--20\% of the data from the respective public benchmarks. 
The specific metrics are detailed as follows:
\textbf{(1) Surround-View 3D Reconstruction:} Adopting the standards from~\cite{wang2025pi,zuo2025dvgt,yang2024depth,wang2025vggt}, we first apply the Umeyama algorithm for scale registration, and subsequently measure Accuracy (Acc) and Completeness (Comp). 
\textbf{(2) Camera Pose Estimation:} Following~\cite{wang2025pi,yang2024depth}, we use Relative Rotation Accuracy (RRA) and Relative Translation Accuracy (RTA) to quantify the relative errors between image pairs. We also report the AUC, defined as the area under the accuracy curve of the minimum between RRA and RTA across various thresholds. 
\textbf{(3) Computational Overhead:} To assess operational efficiency and deployment feasibility, we report latency, parameters count (Params.), and inference speed (FPS).
\textbf{(4) Depth Estimation:} Following~\cite{wang2025pi}, we employ the Absolute Relative error (Abs Rel) and the inlier ratio ($\delta < 1.25$) as our core metrics. 
\textbf{(5) Downstream Applications:} For end-to-end autonomous driving, we evaluate trajectory prediction on the closed-loop Navsim simulator and its Hard benchmark. Following existing works~\cite{du2025sparseworld,zuo2025dvgt}, we report driving metrics including NC, DAC, TTC, Comf., EP, and PDMS. Additionally, for downstream perception tasks such as 3D occupancy prediction on nuScenes, we use mean Average Precision (mAP).


\begin{table*}[!t]
\centering
\small
\caption{\textbf{Quantitative 3D reconstruction results across diverse datasets.} We evaluate the reconstruction quality using Accuracy (Acc) and Completeness (Comp), where lower values indicate better performance ($\downarrow$). To recover the absolute metric scale for a fair comparison, all predicted 3D point maps are aligned with the GT point clouds using the Umeyama~\cite{umeyama1991least} algorithm prior to evaluation.}
\label{tab:quantitative_results_recon}
\resizebox{\textwidth}{!}{
\begin{tabular}{l | c | cc|cc|cc|cc|cc}
\toprule
\multirow{2}{*}{\textbf{Method}} & \multirow{2}{*}{\textbf{Params.}} & \multicolumn{2}{c}{\textbf{Waymo}} & \multicolumn{2}{c}{\textbf{DDAD}} & \multicolumn{2}{c}{\textbf{Lyft}} & \multicolumn{2}{c}{\textbf{PandaSet}} & \multicolumn{2}{c}{\textbf{nuScenes}}  \\
\cmidrule(lr){3-4} \cmidrule(lr){5-6} \cmidrule(lr){7-8} \cmidrule(lr){9-10} \cmidrule(lr){11-12} 
& & Acc $\downarrow$ & Comp $\downarrow$ & Acc $\downarrow$ & Comp $\downarrow$ & Acc $\downarrow$ & Comp $\downarrow$ & Acc $\downarrow$ & Comp $\downarrow$ & Acc $\downarrow$ & Comp $\downarrow$ \\
\midrule
VGGT~\cite{wang2025vggt}       & 1.1B & \textbf{0.389} & 1.565 & 1.413 & 4.010 & 1.660 & 3.789 & 1.402 & 2.082 & 1.191 & 2.417 \\
FastVGGT~\cite{shen2025fastvggt}   & 1.1B & 0.409 & 1.647 & 1.376 & 5.554 & 1.652 & 4.845 & 1.476 & 2.952 & 1.154 & 3.532 \\
LiteVGGT~\cite{shu2025litevggt}   & 1.1B & 0.400 & 1.420 & 1.446 & 3.656 & 1.723 & 4.214 & 1.533 & 2.257 & 1.151 & 1.983 \\
$\pi^3$~\cite{wang2025pi}    & 959M  & 0.494 & \textbf{1.399} & 1.017 & 1.281 & 1.247 & \textbf{1.632} & \textbf{1.238} & \textbf{1.600} & 1.101 & \textbf{1.194} \\
OmniVGGT~\cite{peng2025omnivggt}   & 1.21B & 0.553 & 2.017 & 1.600 & 3.303 & 1.838 & 4.686 & 1.847 & 3.451 & 1.336 & 3.260 \\
DVGT~\cite{zuo2025dvgt}            & 1.73B & 0.493 & 3.283 & 1.172 & 3.718 & \textbf{1.217} & 6.459 & 1.399 & 2.321 & \textbf{1.060} & 2.152 \\
\midrule
\rowcolor{avgblue} 
\textbf{LiAuto-GeoX}   & \textbf{155M}     & 0.547 & 1.729 & \textbf{1.012} & \textbf{1.174} & 1.426 & 2.762 & 1.629 & 2.078 & 1.139 & 1.729 \\
\bottomrule
\end{tabular}
}
\end{table*}

\subsection{Surround-View 3D Reconstruction}

We report quantitative surround-view 3D reconstruction results in Table~\ref{tab:quantitative_results_recon}. Following common reconstruction evaluation protocols, we apply Umeyama~\cite{umeyama1991least} alignment before computing point-cloud accuracy and completeness, which measures the geometric discrepancy between reconstructed point clouds and ground-truth point clouds under aligned scale and pose. 
Across five autonomous driving datasets, LiAuto-GeoX achieves competitive reconstruction quality with only 155M parameters, making it substantially more compact than VGGT-style baselines with over 1B parameters and $\pi^3$ with 959M parameters. 
While large models still achieve stronger accuracy on some datasets, such as Waymo and PandaSet, LiAuto-GeoX maintains comparable point-cloud reconstruction quality under a much smaller model scale. 
Notably, on DDAD, LiAuto-GeoX achieves the best performance in both accuracy and completeness, with an Acc of 1.012 and a Comp of 1.174, outperforming all compared methods including the much larger $\pi^3$. 
On nuScenes, our method also obtains a strong completeness score of 1.729, surpassing VGGT, FastVGGT, LiteVGGT, OmniVGGT, and DVGT. These results suggest that compact driving geometry models can retain competitive reconstruction quality while providing a more practical efficiency profile for deployment.

\begin{figure*}[!t] 
  \centering 
  \includegraphics[width=1.0 \textwidth]{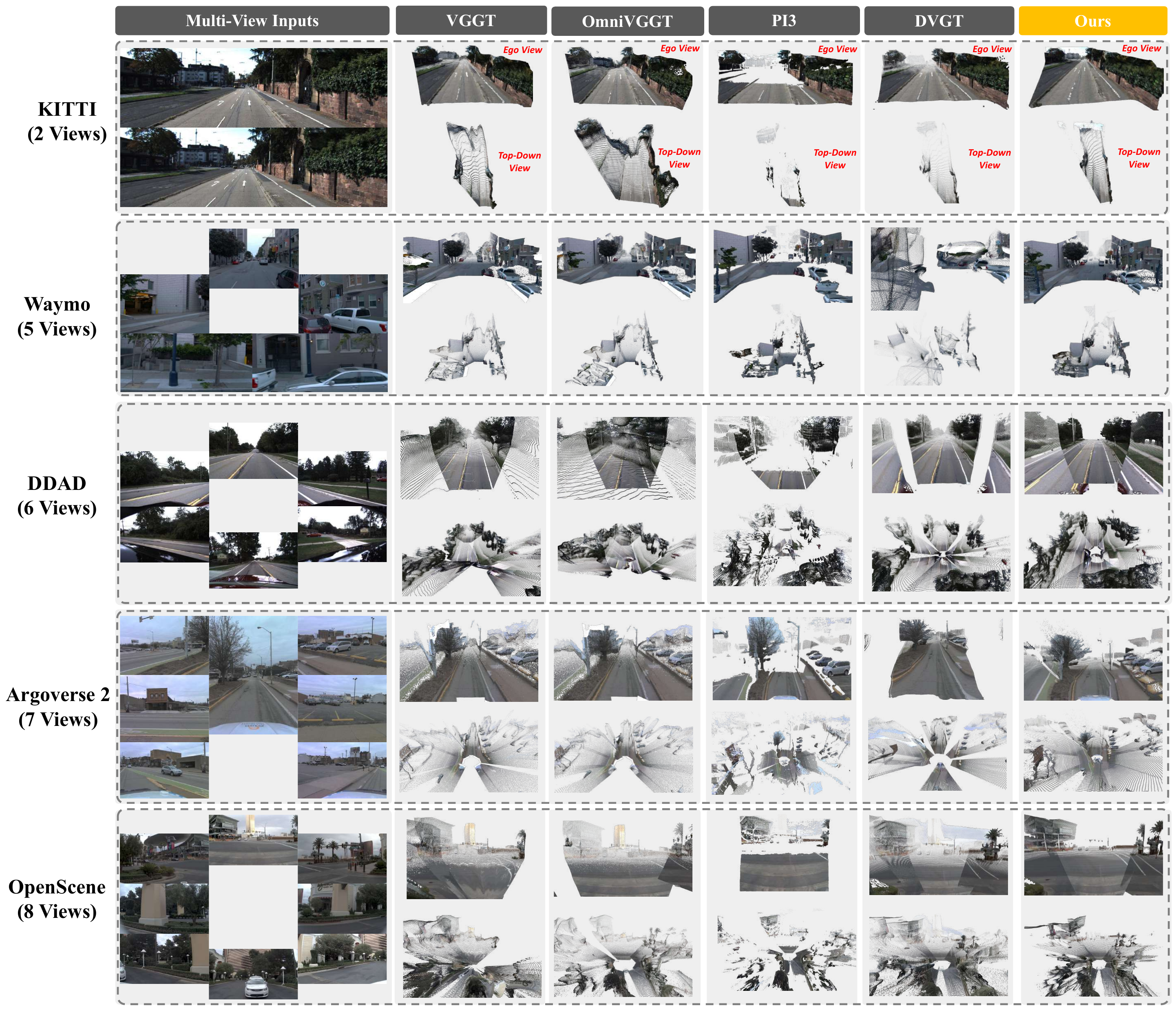} 
  \caption{
   \textbf{Qualitative comparison of multi-view 3D reconstruction.} We compare our method against baselines (VGGT, OmniVGGT, PI3, DVGT) using their default open-source configurations. Each result displays the Ego View (top) and Top-Down View (bottom). Across varying camera counts (2 to 8 views), our model delivers sharper geometry and more complete structures.
  } 
  \label{fig:vis_compare} 
\end{figure*}

We further provide qualitative comparisons in Figure~\ref{fig:vis_compare}. Since Umeyama alignment partially compensates for global scale and pose discrepancies, quantitative metrics alone may understate differences in actual surround-view reconstruction quality. 
The visualization shows that LiAuto-GeoX produces more coherent top-down layouts and more complete road-scene structures across datasets with different camera configurations, including KITTI, Waymo, DDAD, Argoverse 2, and OpenScene. 
Compared with larger generic visual geometry models, which can still exhibit fragmented layouts, inconsistent cross-view geometry, or missing distant structures, our method better preserves the spatial arrangement of roads, vehicles, and scene boundaries in many driving cases. This qualitative evidence complements the aligned point-cloud metrics and highlights the practical reconstruction quality of LiAuto-GeoX under realistic multi-camera driving scenarios.

\begin{table*}[!t]
\centering
\caption{\textbf{Quantitative camera pose estimation results across diverse datasets.} We evaluate the surround-view multi-camera pose estimation performance using Relative Rotation Accuracy (RRA), Relative Translation Accuracy (RTA), and the Area Under the Curve (AUC) of the pose error. For all three metrics, higher values indicate better performance ($\uparrow$). The best results are highlighted in \textbf{bold}.}
\label{tab:pose_estimation}
\resizebox{\textwidth}{!}{
\begin{tabular}{l | c c c | c c c | c c c | c c c}
\toprule
\multirow{2}{*}{\textbf{Method}} & \multicolumn{3}{c}{\textbf{Pandaset}} & \multicolumn{3}{c}{\textbf{DDAD}} & \multicolumn{3}{c}{\textbf{Lyft}} & \multicolumn{3}{c}{\textbf{nuScenes}} \\
\cmidrule(lr){2-4} \cmidrule(lr){5-7} \cmidrule(lr){8-10} \cmidrule(lr){11-13}
& RRA $\uparrow$ & RTA $\uparrow$ & AUC $\uparrow$ & RRA $\uparrow$ & RTA $\uparrow$ & AUC $\uparrow$ & RRA $\uparrow$ & RTA $\uparrow$ & AUC $\uparrow$ & RRA $\uparrow$ & RTA $\uparrow$ & AUC $\uparrow$ \\
\midrule
VGGT~\cite{wang2025vggt}        & 98.66 & 30.66 & 13.35 & 64.49 & 27.50 & 5.928 & 94.41 & 41.25 & 15.32 & 76.50 & 44.00 & 13.19 \\
FastVGGT~\cite{shen2025fastvggt}    & 89.25 & 35.58 & 12.07 & 57.12 & 27.71 & 5.113 & 87.16 & 39.00 & 13.19 & 68.50 & 38.50 & 9.627 \\
LiteVGGT~\cite{shu2025litevggt}    & 98.83 & 33.00 & 15.00 & 94.17 & 38.31 & 14.07 & 99.58 & 39.25 & 18.43 & 97.33 & 40.16 & 16.58  \\
$\pi^3$~\cite{wang2025pi}     & \textbf{100.0} & 38.67 & 21.13 & \textbf{100.0} & 39.36 & 20.38 & \textbf{100.0} & 52.66 & 25.82 & 98.00 & 45.00 & 19.30 \\
OmniVGGT~\cite{peng2025omnivggt}     & 86.00 & 34.41 & 11.20 & 55.08 & 20.35 & 3.957 & 77.25 & 33.53 & 7.334 & 48.00 & 32.66 & 5.027 \\
\midrule
\rowcolor{avgblue}
\textbf{LiAuto-GeoX}    & \textbf{100.0} & \textbf{100.0} & \textbf{91.24} & \textbf{100.0} & \textbf{99.92} & \textbf{94.94} & \textbf{100.0} & \textbf{98.75} & \textbf{92.14} & \textbf{100.0} & \textbf{98.33} & \textbf{92.17} \\
\bottomrule
\end{tabular}
}
\end{table*}

\begin{table*}[!t]
\centering
\caption{\textbf{Comparison of inference efficiency across diverse datasets.} 
We report inference latency (ms), processing speed in frames per second (FPS), and GPU memory consumption (GB) for each dataset. Lower latency and memory usage ($\downarrow$), and higher FPS ($\uparrow$) indicate better efficiency. All measurements are evaluated on a single A100 (80GB) GPU. The best results are highlighted in \textbf{bold}.}
\label{tab:time_effcient}
\resizebox{\textwidth}{!}{
\begin{tabular}{l | c c c | c c c | c c c | c c c}
\toprule
\multirow{2}{*}{\textbf{Method}} 
& \multicolumn{3}{c}{\textbf{Waymo}} 
& \multicolumn{3}{c}{\textbf{DDAD}} 
& \multicolumn{3}{c}{\textbf{Argoverse 2}} 
& \multicolumn{3}{c}{\textbf{OpenScene}} \\
\cmidrule(lr){2-4} 
\cmidrule(lr){5-7} 
\cmidrule(lr){8-10} 
\cmidrule(lr){11-13}
& Latency $\downarrow$ & FPS $\uparrow$ & Mem $\downarrow$
& Latency $\downarrow$ & FPS $\uparrow$ & Mem $\downarrow$
& Latency $\downarrow$ & FPS $\uparrow$ & Mem $\downarrow$
& Latency $\downarrow$ & FPS $\uparrow$ & Mem $\downarrow$ \\
\midrule
VGGT~\cite{wang2025vggt}            
& 102.84 & 9.72 & 7.812 
& 116.10 & 8.61 & 8.042 
& 144.92 & 6.90 & 8.813 
& 129.50 & 7.72 & 8.431 \\

FastVGGT~\cite{shen2025fastvggt}        
& 59.65 & 16.76 & 5.387 
& 67.34 & 14.85 & 5.596 
& 84.05 & 11.90 & 6.128 
& 75.11 & 13.31 & 5.841 \\

LiteVGGT~\cite{shu2025litevggt}        
& 63.42 & 15.77 & 4.781 
& 70.29 & 14.23 & 4.942 
& 87.12 & 11.48 & 5.375 
& 78.44 & 12.75 & 5.120 \\

$\pi^3$~\cite{wang2025pi}         
& 82.10 & 12.18 & 3.791 
& 85.86 & 11.65 & 3.809 
& 116.38 & 8.59 & 3.896 
& 100.26 & 9.97 & 3.854 \\

OmniVGGT~\cite{peng2025omnivggt}        
& 106.77 & 9.37 & 6.419 
& 118.43 & 8.44 & 6.596 
& 154.78 & 6.46 & 7.428 
& 132.53 & 7.55 & 7.012 \\

DVGT~\cite{zuo2025dvgt}                 
& 110.89 & 9.02 & 8.340 
& 124.59 & 8.03 & 8.596 
& 155.04 & 6.45 & 9.414 
& 138.16 & 7.24 & 8.977 \\
\midrule
\rowcolor{avgblue}
\textbf{LiAuto-GeoX }       
& \textbf{50.22} & \textbf{19.91} & \textbf{1.752} 
& \textbf{58.83} & \textbf{17.00} & \textbf{1.855} 
& \textbf{63.41} & \textbf{15.77} & \textbf{2.367} 
& \textbf{58.21} & \textbf{17.18} & \textbf{2.100} \\
\bottomrule
\end{tabular}
}
\end{table*}

\subsection{Camera Pose Estimation}
Table~\ref{tab:pose_estimation} presents the camera pose estimation results.
In this evaluation, the model takes surround-view multi-camera images as input to predict camera poses, and the reported metrics are averaged across all views.
LiAuto-GeoX achieves the strongest overall performance across all evaluated datasets.
In particular, our method reaches 100.0 rotation accuracy on PandaSet, DDAD, Lyft, and nuScenes, while also achieving near-perfect translation accuracy, with 100.0 on PandaSet and 99.92 on DDAD.
The gains are more pronounced in AUC: LiAuto-GeoX improves from 21.13 to 91.24 on PandaSet, from 20.38 to 94.94 on DDAD, from 25.82 to 92.14 on Lyft, and from 19.30 to 92.17 on nuScenes compared with the strongest baseline $\pi^3$.
These results demonstrate that LiAuto-GeoX is well suited for surround-view pose estimation, where camera rigs exhibit a structured and physically constrained spatial layout.
Compared with generic 3D foundation models, LiAuto-GeoX produces more stable and coherent cross-camera poses, which are essential for reliable 3D reasoning in autonomous driving.

\subsection{Inference Efficiency Comparison}

We compare the inference efficiency of LiAuto-GeoX with recent visual geometry models in Table~\ref{tab:time_effcient}. To reflect realistic driving inputs, each dataset is evaluated with its native surround-view camera setup: 5 cameras for Waymo, 6 for DDAD, 7 for Argoverse 2, and 8 for OpenScene. All methods are measured on driving sequences with $N=20$ frames using a single A100 GPU, and we report latency, FPS, and GPU memory consumption.

LiAuto-GeoX consistently achieves the best efficiency across all datasets. It reaches 50.22 ms latency and 19.91 FPS on Waymo, and maintains stable performance as the camera number increases, achieving 63.41 ms on Argoverse 2 and 58.21 ms on OpenScene. In contrast, large visual geometry models such as VGGT, OmniVGGT, and DVGT require substantially higher latency and memory, with DVGT reaching 155.04 ms and 9.414 GB on Argoverse 2. LiAuto-GeoX only consumes 1.752--2.367 GB memory across datasets, showing a clear latency-memory advantage under realistic multi-camera sequence inputs. These results validate the effectiveness of compact geometry transfer for real-time onboard dense 3D reconstruction.

\begin{table*}[!t]
\centering
\caption{\textbf{Quantitative depth results across diverse datasets.}
We evaluate depth estimation in two driving scenarios: surround-view multi-camera monocular depth estimation on large-scale driving datasets, and video depth estimation on KITTI.
All results are evaluated under the same scale, and FPS is measured on the same single GPU for fair efficiency comparison.}
\label{tab:monocular_video_depth}
\resizebox{\textwidth}{!}{
\begin{tabular}{l | c c | c c | c c | c c | c c c}
\toprule
\multirow{2}{*}{\textbf{Method}} & \multicolumn{2}{c}{\textbf{Lyft}} & \multicolumn{2}{c}{\textbf{Pandaset}} & \multicolumn{2}{c}{\textbf{Waymo}} & \multicolumn{2}{c}{\textbf{nuScenes}} & \multicolumn{3}{c}{\textbf{KITTI}} \\
\cmidrule(lr){2-3} \cmidrule(lr){4-5} \cmidrule(lr){6-7} \cmidrule(lr){8-9} \cmidrule(lr){10-12}
& Abs Rel $\downarrow$ & $\delta<1.25 \uparrow$ & Abs Rel $\downarrow$ & $\delta<1.25 \uparrow$ & Abs Rel $\downarrow$ & $\delta<1.25 \uparrow$ & Abs Rel $\downarrow$ & $\delta<1.25 \uparrow$ & Abs Rel $\downarrow$ & $\delta<1.25 \uparrow$ & FPS $\uparrow$ \\
\midrule
VGGT~\cite{wang2025vggt}        & 0.158 & 0.801 & 0.243 & 0.729 & 0.176 & 0.811 & 0.337 & 0.579 & 0.058 & 0.961 & 84.46 \\
$\pi^3$~\cite{wang2025pi}       & 0.352 & 0.468 & 0.252 & 0.651 & 0.187 & 0.747 & 0.230 & 0.634 & \textbf{0.039} & \textbf{0.985} & 110.2 \\
OmniVGGT~\cite{peng2025omnivggt}& 0.270 & 0.622 & 0.241 & 0.658 & 0.281 & 0.623 & 0.269 & 0.618 & 0.053 & 0.964 & 84.18 \\
DVGT~\cite{zuo2025dvgt}         & 0.224 & 0.685 & 0.268 & 0.635 & 0.148 & 0.832 & 0.218 & 0.648 & 0.109 & 0.894 & 77.80 \\
\midrule
\rowcolor{avgblue}
\textbf{LiAuto-GeoX}      & \textbf{0.136} & \textbf{0.795} & \textbf{0.104} & \textbf{0.876} & \textbf{0.114} & \textbf{0.856} & \textbf{0.115} & \textbf{0.873} & 0.084 & 0.931 & \textbf{223.8} \\
\bottomrule
\end{tabular}
}
\end{table*}

\subsection{Depth Estimation}
As shown in Table~\ref{tab:monocular_video_depth}, LiAuto-GeoX achieves the best overall performance on surround-view multi-camera driving datasets.
On Lyft, PandaSet, Waymo, and nuScenes, it obtains the lowest Abs Rel among all methods, reaching 0.136, 0.104, 0.114, and 0.115, respectively.
It also achieves the best ($\delta < 1.25$) on PandaSet, Waymo, and nuScenes.
These results show that LiAuto-GeoX generalizes well across diverse driving platforms and camera configurations, where side and rear views often involve larger viewpoint changes, weaker overlap, and long-range structures.

Compared with general-purpose visual geometry models, LiAuto-GeoX is more robust in surround-view driving scenarios.
VGGT and $\pi^3$ perform strongly on KITTI, but their performance drops on large-scale surround-view datasets, such as nuScenes, Lyft, and PandaSet.
This suggests that broad visual geometry pretraining does not necessarily transfer optimally to ego-centric multi-camera driving scenes.
In contrast, LiAuto-GeoX is explicitly trained and distilled for driving geometry, leading to more consistent depth quality across datasets.

For KITTI, we follow the video-depth evaluation setting.
Since KITTI mainly contains forward-facing camera sequences, it is closer to the dominant setting of many generic depth and visual geometry pretraining pipelines, giving methods such as $\pi^3$ and VGGT a natural advantage.
Although $\pi^3$ achieves the best KITTI accuracy with 0.039 Abs Rel and 0.985 ($\delta < 1.25$), LiAuto-GeoX remains competitive while being substantially faster, reaching 223.8 FPS on a single GPU.
Overall, LiAuto-GeoX provides a stronger trade-off between depth accuracy, cross-dataset robustness, and real-time efficiency for onboard driving deployment.

\subsection{Downstream Task Comparison}
\textbf{Closed-loop planning.} We evaluate the transferability of LiAuto-GeoX on closed-loop planning using the NAVSIM~\cite{dauner2024navsim} v1 NAVTEST split, with results summarized in Table~\ref{tab:navsim_v1_traj}. 
For this downstream evaluation, we freeze the backbone parameters learned from geometric reconstruction to extract spatial features. We then train only a lightweight planning decoder on NAVSIM in a supervised manner to predict future trajectories. Because the backbone is kept strictly frozen without additional fine-tuning or reinforcement learning, this setting properly evaluates whether reconstructed 3D geometry can serve as a transferable representation for planning, rather than relying on task-specific adaptation of the entire network.

LiAuto-GeoX achieves 90.6 PDMS using camera input only, outperforming prior geometric reconstruction methods such as DVGT-2 and DVGT-2-NAVSIM, and also surpassing representative end-to-end, VLA, and world-model-based methods. Notably, our method improves over DVGT-2-NAVSIM from 90.3 to 90.6 PDMS, while maintaining strong EP performance of 85.8. These results indicate that the dense geometry learned by LiAuto-GeoX provides useful spatial cues for downstream trajectory prediction and closed-loop planning, even without task-specific backbone tuning or RL-based score boosting.

\begin{table*}[!t]
\tiny
\centering
\caption{\textbf{Closed-loop planning results on NAVSIM v1 NAVTEST split.} Future states represent world-modeling-based methods. C and L denote camera and LiDAR, respectively.}
\label{tab:navsim_v1_traj}
\resizebox{\textwidth}{!}{
\begin{tabular}{l | c | c c c c c >{\columncolor{avgblue}}c}
\toprule
Method & Input & NC $\uparrow$ & DAC $\uparrow$ & TTC $\uparrow$ & Comf. $\uparrow$ & EP $\uparrow$ & PDMS $\uparrow$ \\
\midrule
\multicolumn{8}{l}{\textit{Traditional End-to-End Methods}} \\
UniAD~\cite{hu2023planning} & C & 97.8 & 91.9 & 92.9 & \textbf{100} & 78.8 & 83.4 \\
Transfuser~\cite{prakash2021multi} & C \& L & 97.7 & 92.8 & 92.8 & \textbf{100} & 79.2 & 84.0 \\
Hydra-MDP~\cite{li2024hydra} & C \& L & \textbf{98.3} & 96.0 & 94.6 & \textbf{100} & 78.7 & 86.5 \\
GoalFlow~\cite{xing2025goalflow} & C \& L & \textbf{98.3} & 93.8 & 94.3 & \textbf{100} & 79.8 & 85.7 \\
ARTEMIS~\cite{feng2025artemis} & C \& L & \textbf{98.3} & 95.1 & 94.3 & \textbf{100} & 81.4 & 87.0 \\
DiffusionDrive~\cite{liao2025ddrive} & C \& L & 98.2 & 96.2 & 94.7 & \textbf{100} & 82.2 & 88.1 \\
VADv2~\cite{chen2024vadv2} & C & \textbf{98.3} & \textbf{97.4} & \textbf{95.7} & \textbf{100} & 82.3 & 89.3 \\
DriveSuprim~\cite{yao2026drivesuprim} & C \& L & 97.8 & 97.3 & 93.6 & \textbf{100} & \textbf{86.7} & \textbf{89.9} \\
\midrule
\multicolumn{8}{l}{\textit{Vision-Language-Action Methods}} \\
AdaThinkDrive~\cite{luo2025adathinkdrive} & C & 98.5 & 94.4 & 94.9 & \textbf{100} & 79.9 & 86.2 \\
AutoVLA~\cite{zhou2025autovla} & C & 98.4 & 95.6 & \textbf{98.0} & 99.9 & \textbf{85.9} & 89.1 \\
ReCogDrive~\cite{li2025recogdrive} & C & 98.2 & 97.8 & 95.2 & 99.8 & 83.5 & 89.6 \\
DriveVLA-W0~\cite{li2025drivevla} & C & \textbf{98.7} & \textbf{99.1} & 95.3 & 99.3 & 83.3 & \textbf{90.2} \\
\midrule
\multicolumn{8}{l}{\textit{World Model Methods}} \\
DrivingGPT~\cite{chen2025drivinggpt} & C & 98.9 & 90.7 & 94.9 & 95.6 & 79.7 & 82.4 \\
LAW~\cite{li2025enhancing} & C & 96.4 & 95.4 & 88.7 & 99.9 & 81.7 & 84.6 \\
Epona~\cite{zhang2025epona} & C & 97.9 & 95.1 & 93.8 & 99.9 & 80.4 & 86.2 \\
WoTE~\cite{li2025end} & C \& L & 98.5 & 96.8 & 94.9 & 99.9 & \textbf{81.9} & 88.3 \\
PWM~\cite{zhao2026forecasting} & C & 98.6 & 95.9 & 95.4 & \textbf{100} & 81.8 & 88.1 \\
DriveLaW~\cite{xia2026drivelaw} & C & \textbf{99.0} & \textbf{97.1} & \textbf{96.7} & \textbf{100} & 81.3 & \textbf{89.1} \\
\midrule
\multicolumn{8}{l}{\textit{Geometric Reconstruction Methods}} \\
DVGT-2~\cite{zuo2026dvgt} & C & 97.8 & 97.2 & 93.9 & \textbf{100} & 83.4 & 88.6 \\
DVGT-2-NAVSIM~\cite{zuo2026dvgt} & C & \textbf{98.7} & \textbf{97.9} & \textbf{95.8} & \textbf{100} & 84.3 & 90.3 \\
\textbf{LiAuto-GeoX} & C & 98.6 & 97.5 & 95.2 & \textbf{100} & \textbf{85.8} & \textbf{90.6} \\
\bottomrule
\end{tabular}
}
\end{table*}

\begin{table*}[!t]
\tiny
\centering
\caption{\textbf{4D OCC forecasting performance on Occ3D-nuScenes}~\cite{tian2023occ3d}.
All methods use camera inputs for forecasting. We report semantic mIoU and geometric IoU at future horizons of 1s, 2s, and 3s. Avg. denotes the average performance over the three horizons.}
\label{tab:occ_world_model}
\resizebox{\textwidth}{!}{
\begin{tabular}{l | c c c >{\columncolor{avgblue}}c | c c c >{\columncolor{avgblue}}c}
\toprule
\multicolumn{1}{c|}{\multirow{2}{*}{Method}} & \multicolumn{4}{c|}{Semantic mIoU (\%) $\uparrow$} & \multicolumn{4}{c}{Geometric IoU (\%) $\uparrow$} \\
\cmidrule{2-5} \cmidrule{6-9}
\multicolumn{1}{c|}{} & 1s & 2s & 3s & Avg. & 1s & 2s & 3s & Avg. \\
\midrule
PreWorld~\cite{li2025semi} & 12.27 & 9.24 & 7.15 & 9.55 & 23.62 & 21.62 & 19.63 & 21.62 \\
Occ-LLM~\cite{xu2025occ} & 11.28 & 10.21 & 9.13 & 10.21 & 27.11 & 24.07 & 20.19 & 23.79 \\
DFIT-OccWorld~\cite{zhang2024efficient} & 13.38 & 10.16 & 7.96 & 10.50 & 19.18 & 16.85 & 25.02 & 17.02 \\
OccTENS-F~\cite{jin2026occtens} & 17.17 & 10.38 & 7.82 & 11.79 & 27.60 & 25.14 & 20.33 & 24.35 \\
SparseWorld~\cite{dang2026sparseworld} & 14.93 & 13.15 & 11.51 & 13.20 & 22.96 & 22.10 & 21.05 & 22.03 \\
DOME-STC~\cite{gu2024dome} & 17.79 & 14.23 & 11.58 & 14.53 & 26.39 & 23.20 & 20.42 & 23.33 \\
I$^2$-World-STC~\cite{liao2025i2} & 21.67 & 18.78 & 16.47 & 18.97 & 30.55 & 28.76 & 26.99 & 28.77 \\
DOME-F~\cite{gu2024dome} & 24.12 & 17.41 & 13.24 & 18.25 & 35.18 & 27.90 & 23.44 & 28.84 \\
DTT~\cite{xu2025delta} & 24.87 & 18.30 & 15.63 & 19.60 & 38.98 & 37.45 & 31.89 & 36.11 \\
COME~\cite{shi2026come} & 26.56 & 21.73 & 18.49 & 22.26 & 48.08 & 43.84 & 40.28 & 44.07 \\
SparseWorld-TC~\cite{du2025sparseworld} & 26.33 & 24.00 & 22.05 & 24.13 & 48.20 & 46.88 & 45.13 & 46.64 \\
\textbf{LiAuto-GeoX} & \textbf{26.75} & \textbf{24.47} & \textbf{22.58} & \textbf{24.63} & \textbf{49.09} & \textbf{47.81} & \textbf{46.12} & \textbf{47.67} \\
\bottomrule
\end{tabular}
}
\end{table*}

\textbf{4D occupancy forecasting.}
We evaluate LiAuto-GeoX on camera-based 4D occupancy forecasting using Occ3D-nuScenes, with results shown in Table~\ref{tab:occ_world_model}. LiAuto-GeoX uses DINOv2-Small as its reconstruction backbone, whose parameter scale is comparable to the ResNet-34 encoder in the strongest baseline SparseWorld-TC. This makes the comparison primarily reflect the quality of the learned geometric representation rather than encoder capacity. LiAuto-GeoX achieves the best overall performance, reaching 24.63\% average semantic mIoU and 47.67\% average geometric IoU.

Compared with SparseWorld-TC, LiAuto-GeoX improves long-horizon forecasting at 3s from 22.05\% to 22.58\% in semantic mIoU and from 45.13\% to 46.12\% in geometric IoU. The consistent gains over future horizons indicate that the reconstructed dense geometry provides useful spatial cues for predicting future occupancy, supporting its transferability beyond current-frame reconstruction.

\begin{table}[!t]
\tiny
\centering
\caption{\textbf{Ablation study of the proposed distillation strategies.} We evaluate the effectiveness of different distillation components on a sampled subset of the Waymo dataset. The results of the teacher model (upper bound) and the best-performing student model are highlighted in \textbf{bold}.}
\label{tab:distill_ablation}
\resizebox{\linewidth}{!}{
\begin{tabular}{l | c c c | c c c | c c}
\toprule
\multirow{2}{*}{\textbf{Setting}} & \multicolumn{3}{c|}{\textbf{3D Reconstruction}} & \multicolumn{3}{c|}{\textbf{Camera Pose}} & \multicolumn{2}{c}{\textbf{Depth}} \\
\cmidrule(lr){2-4} \cmidrule(lr){5-7} \cmidrule(lr){8-9}
& Acc $\downarrow$ & Comp $\downarrow$ & NC $\uparrow$ & RRA $\uparrow$ & RTA $\uparrow$ & AUC $\uparrow$ & Abs Rel $\downarrow$ & $\delta < 1.25 \uparrow$ \\
\midrule
\textbf{(a) Teacher Model}        & \textbf{0.483} & \textbf{1.313} & \textbf{0.674} & \textbf{100.0} & \textbf{98.33} & \textbf{84.24} & \textbf{0.062} & \textbf{0.957} \\
\textbf{(b) Vanilla}              & 0.812 & 2.451 & 0.482 & 91.25 & 42.18 & 61.32 & 0.145 & 0.763 \\

\textbf{(c) Logit Distillation}   & 0.735 & 2.184 & 0.515 & 94.60 & 58.45 & 66.85 & 0.118 & 0.812 \\
\midrule
\multicolumn{9}{l}{\textit{w/ Geometric Information}} \\
\midrule
\textbf{(d) Mask-Guided}          & 0.621 & 1.949 & 0.628 & 95.80 & 64.20 & 69.45 & 0.082 & 0.895 \\
\textbf{(e) Relation Distillation}& 0.684 & 2.012 & 0.548 & 99.10 & 88.65 & 77.20 & 0.105 & 0.834 \\
\rowcolor{avgblue}
\textbf{(f) Ours}                 & \textbf{0.614} & \textbf{1.871} & \textbf{0.638} & \textbf{100.0} & \textbf{95.51} & \textbf{79.71} & \textbf{0.075} & \textbf{0.914} \\
\bottomrule
\end{tabular}
}
\end{table}

\begin{table}[t]
\tiny
\centering
\caption{\textbf{Effect of Token Masking Strategy.} We study different activation-based token masking strategies and evaluate their impact on depth estimation and 3D reconstruction, especially for distant and geometrically informative regions. The best results are highlighted in \textbf{bold}.}
\label{tab:masking_ablation}
\resizebox{\linewidth}{!}{
\begin{tabular}{l | c c c | c c c | c c}
\toprule
\multirow{2}{*}{\textbf{Masking Strategy}} & \multicolumn{3}{c|}{\textbf{3D Reconstruction}} & \multicolumn{3}{c|}{\textbf{Camera Pose}} & \multicolumn{2}{c}{\textbf{Depth}} \\
\cmidrule(lr){2-4} \cmidrule(lr){5-7} \cmidrule(lr){8-9}
& Acc $\downarrow$ & Comp $\downarrow$ & NC $\uparrow$ & RRA $\uparrow$ & RTA $\uparrow$ & AUC $\uparrow$ & Abs Rel $\downarrow$ & $\delta < 1.25 \uparrow$ \\
\midrule
\textbf{(a) Random Masking}             & 0.655 & 1.962 & 0.608 & 96.85 & 85.40 & 76.15 & 0.088 & 0.886 \\
\textbf{(b) Low Act. (Act. $\le$ Avg)}  & 0.672 & 2.015 & 0.595 & 96.10 & 83.75 & 75.30 & 0.094 & 0.872 \\
\textbf{(c) Top-50\% High Act.}         & 0.628 & 1.895 & 0.625 & 97.92 & 87.85 & 77.62 & 0.079 & 0.905 \\
\rowcolor{avgblue}
\textbf{(d) High Act. (Act. $>$ Avg)}   & \textbf{0.614} & \textbf{1.871} & \textbf{0.638} & \textbf{98.51} & \textbf{89.20} & \textbf{78.45} & \textbf{0.075} & \textbf{0.914} \\
\bottomrule
\end{tabular}
}
\end{table}

\subsection{Ablation Study}

\textbf{Effect of Geometry-Preserving Distillation.}
Table~\ref{tab:distill_ablation} ablates the proposed distillation strategies on a sampled subset of Waymo.
Setting (a) is our self-trained 1.1B-parameter teacher model, which serves as the upper bound.
Setting (b) trains the compact student from scratch without teacher guidance, leading to clear degradation across all metrics, especially in camera pose estimation and 3D reconstruction.
Setting (c) applies conventional logit distillation from the teacher, which improves over vanilla training but remains limited, indicating that output-level supervision alone is insufficient to transfer dense driving geometry.

Introducing geometric information brings more targeted improvements.
Mask-guided depth-aware distillation in (d) substantially improves reconstruction and depth estimation, reducing Acc from 0.735 to 0.621 and Abs Rel from 0.118 to 0.082 compared with logit distillation.
Relative-pose relational distillation in (e) mainly benefits cross-view geometry, increasing RRA from 94.60 to 99.10, RTA from 58.45 to 88.65, and AUC from 66.85 to 77.20.
Combining both strategies in (f) achieves the best student performance, with 0.614 Acc, 1.871 Comp, 0.638 NC, 100.0 RRA, 95.51 RTA, 79.71 AUC, 0.075 Abs Rel, and 0.914 ($\delta < 1.25$).
These results show that the two geometry-preserving objectives are complementary: mask-guided depth-aware distillation improves local metric fidelity, while relative-pose relational distillation strengthens global cross-view consistency.

\textbf{Effect of Token Masking Strategy.}
Table~\ref{tab:masking_ablation} compares different activation-based masking strategies for depth-aware conditioning.
Random masking provides limited guidance, while selecting low-activation tokens leads to worse reconstruction and depth performance, suggesting that low-response regions are less informative for geometric transfer.
In contrast, focusing on high-activation tokens consistently improves all metrics.
The Top-50\% high-activation strategy reduces Abs Rel from 0.088 to 0.079 compared with random masking and improves NC from 0.608 to 0.625.

Using tokens with activations above the average achieves the best overall performance, reaching 0.614 Acc, 1.871 Comp, 0.638 NC, 0.075 Abs Rel, and 0.914 ($\delta < 1.25$).
It also improves camera pose consistency, achieving 98.51 RRA, 89.20 RTA, and 78.45 AUC.
These results indicate that teacher activations provide an effective signal for identifying geometrically informative regions, and that conditioning the student on high-response tokens better preserves depth fidelity, 3D reconstruction quality, and cross-view consistency.

\begin{figure*}[t] 
  \centering 
  \includegraphics[width=1.0 \textwidth]{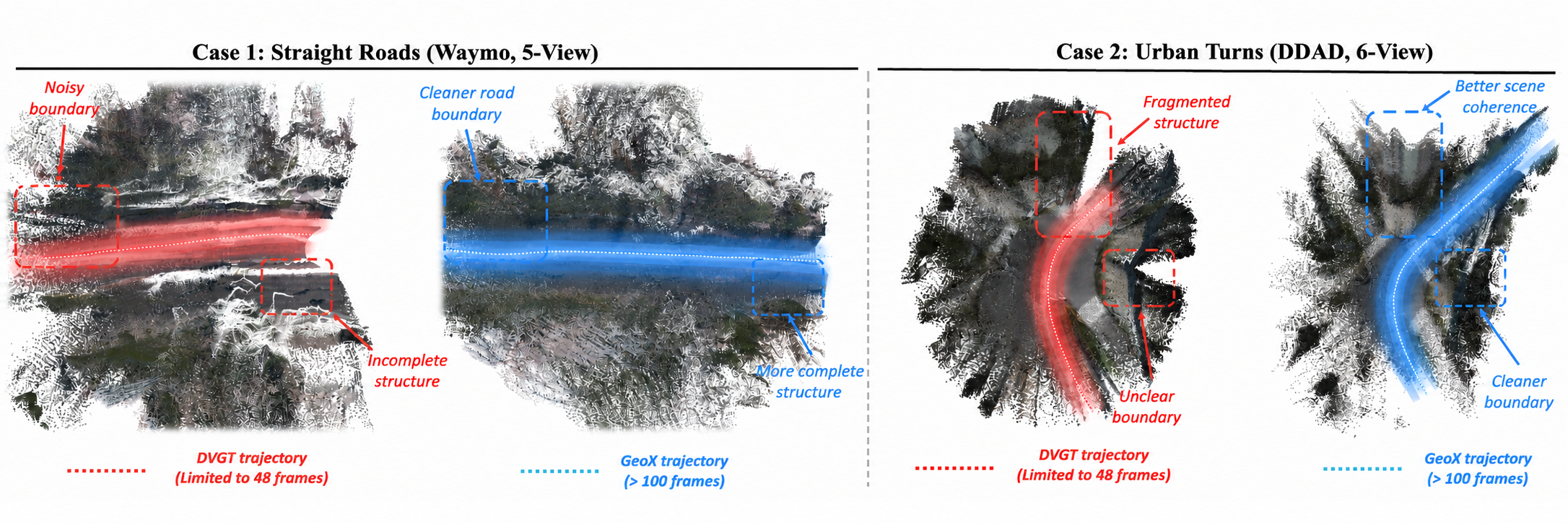} 
    \caption{\textbf{Qualitative comparison of surround-view reconstruction.}
    We compare LiAuto-GeoX with DVGT on Waymo and DDAD sequences. Red and blue trajectories denote DVGT and our method, respectively. LiAuto-GeoX reconstructs cleaner road boundaries, more complete structures, and more coherent top-down geometry across different camera configurations.}
  \label{fig:qualitative_seq} 
\end{figure*}

\begin{figure*}[t] 
  \centering 
  \includegraphics[width=1.0 \textwidth]{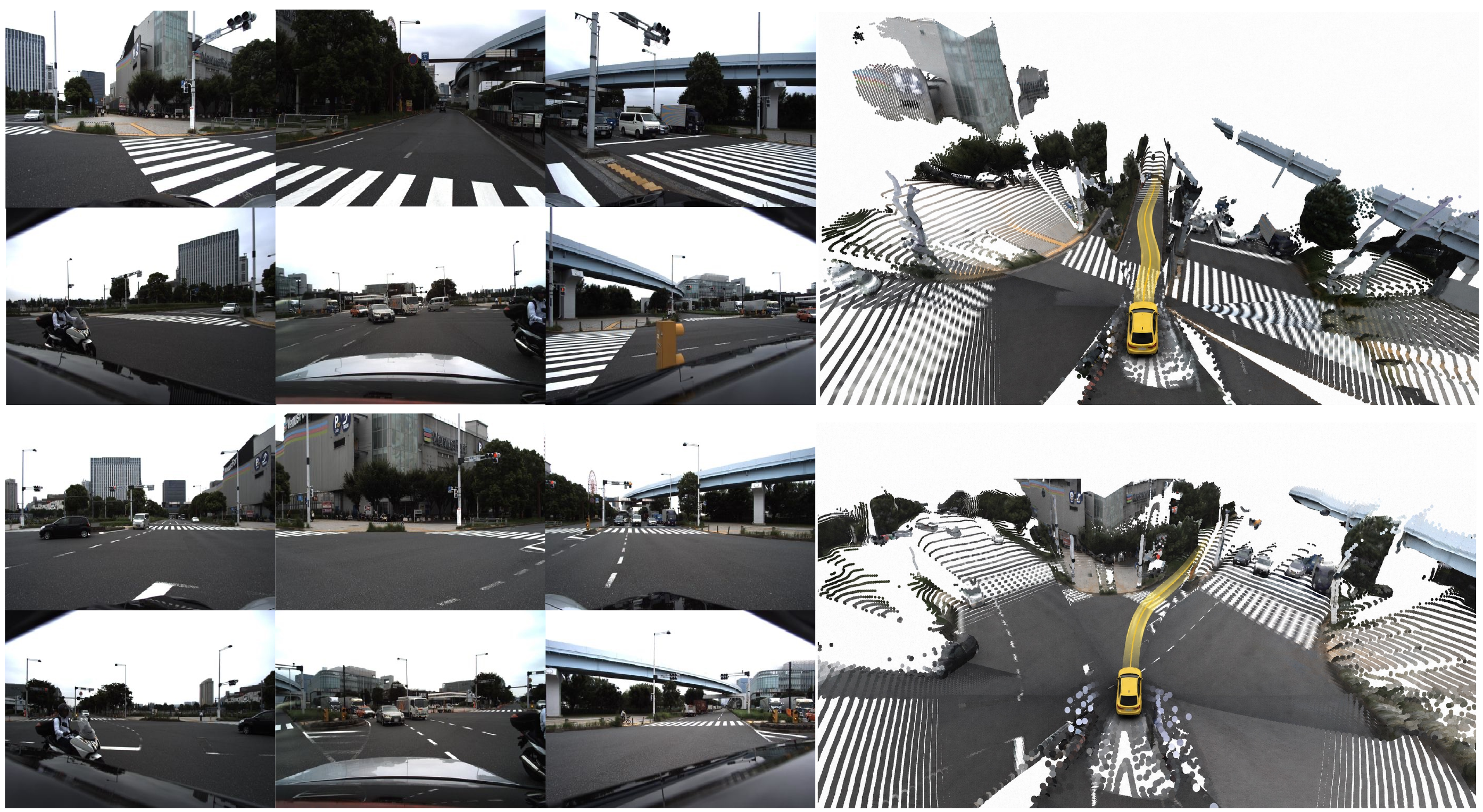} 
\caption{\textbf{Qualitative visualizations.} Surround-view inputs and reconstructed 3D driving scenes with driving trajectories.}
  \label{fig:qualitative_seq_ddad} 
\end{figure*}

\subsection{Qualitative Analysis}
Figure~\ref{fig:qualitative_seq} compares LiAuto-GeoX with DVGT on representative Waymo and DDAD driving sequences. DVGT often produces fragmented structures and noisy road boundaries in regions with limited cross-view overlap. In contrast, LiAuto-GeoX reconstructs more coherent top-down geometry, with clearer road layouts, sharper scene boundaries, and more complete surrounding structures, indicating better cross-view spatial consistency.

The difference is also reflected in sequence-level scalability. Since DVGT is a 1.7B-parameter model, inference on long surround-view sequences such as $50\times5$ Waymo frames or $50\times6$ DDAD frames already approaches the practical compute limit. By contrast, the compact LiAuto-GeoX model can continue reconstruction over sequences of 100 frames or more, as evidenced by the longer blue trajectory in the Waymo visualization. This enables more stable long-sequence visualization and makes dense 3D reconstruction more practical for real driving scenarios.

Additional visualizations in Figure~\ref{fig:qualitative_seq_ddad} further highlight the ability of LiAuto-GeoX to transform surround-view multi-camera observations into structured 3D driving geometry.
The left side presents synchronized multi-view inputs, while the right side shows the reconstructed scenes together with driving trajectories.
Across intersections and turning scenarios, LiAuto-GeoX preserves coherent road topology, continuous drivable-space layout, and stable trajectory-aware geometry.
The reconstructed scenes remain well organized around crosswalks, road boundaries, and nearby static structures, indicating that the learned representation goes beyond isolated depth prediction and provides a structured geometric basis for driving-scene understanding.

\section{Conclusion}

In this work, we have presented LiAuto-GeoX, an efficient grounded driving transformer that has revisited dense 3D reconstruction as a deployable geometric representation for autonomous driving. 
Rather than treating reconstruction as an offline perception objective, we have focued on the central deployment question: whether dense visual geometry can be made accurate, spatially consistent, and sufficiently efficient for onboard driving systems. 
To this end, we have first constructed a large-scale driving geometry teacher from surround-view driving data, and have then transferred its capability to a compact 155M-parameter student through geometry-preserving distillation.
The proposed mask-guided depth-aware distillation has preserved fine-grained metric structures in geometrically informative regions, while relative-pose relational distillation has enforced cross-view consistency under surround-view camera geometry.

Extensive experiments have demonstrated that LiAuto-GeoX has achieved a strong balance between reconstruction fidelity and real-time efficiency, running at 220 FPS on KITTI video sequences while maintaining high-quality dense geometry.
More importantly, the learned representation has transferred effectively beyond reconstruction across diverse driving tasks, improving trajectory prediction, occupancy prediction, and future-frame prediction.
These results have suggested that efficient dense 3D reconstruction can serve as more than a visual geometry task: it can become a scalable geometric substrate for downstream autonomy.
By bridging high-capacity geometry learning and compact onboard deployment, LiAuto-GeoX has highlighted a practical path toward deployable, geometry-grounded representations for real-world driving intelligence.


\section*{Author List}

\textbf{Core Contributors (Equal contribution):} Jiawei Lian, Haoyi Sun, Yang Wu, Lifu Mu

\textbf{Contributors:} Siyuan Wang, Jiawei Chen, Feng Gu, Shanshan Li, Jing Luo, Hao Ma, Xuecheng Ouyang, Xuhao Pan, Xudong Rao, Zichao Shen, Qian Wang, Hongfu Yang, Hongyuan Zhang, Pengfei Yu, Ren Chang

\textbf{Technique Leaders:} Ning Mao, Tao Wei

\textbf{Company Supervisors:} Pan Zhou, Kun Zhan

\textbf{Supervisors:} Le Hui, Jian Yang

\newpage
\appendix
\bibliographystyle{plainnat}
\bibliography{main}

\end{document}